\documentclass[letterpaper, 10 pt, conference]{ieeeconf}  

\IEEEoverridecommandlockouts                              

\usepackage{amsmath} 
\usepackage{amssymb}  
\usepackage{tabularx}
\usepackage{array}
\usepackage{xcolor}
\usepackage{cite}
\usepackage[pagebackref,breaklinks,colorlinks=true,citecolor=red, urlcolor=magenta]{hyperref}
\usepackage{bm}
\usepackage{graphicx}
\usepackage{booktabs}
\usepackage{capt-of}
\usepackage{xspace}
\usepackage{xcolor}
\usepackage{subcaption}
\definecolor{linkblue}{rgb}{0,0,0.8}

\title{\LARGE \bf
GentleHumanoid: Learning Upper-body Compliance for Contact-rich  Human and Object Interaction}

\author{Qingzhou Lu$^*$, Yao Feng$^*$, Baiyu Shi, Michael Piseno, Zhenan Bao, C. Karen Liu \\
\textbf{Stanford University} 
}

\newcommand*{\modelnametext}{GentleHumanoid\xspace}
\newcommand*{\modelname}{{GentleHumanoid}\xspace}
\newcommand*{\baselineA}{{Vanilla-RL}\xspace}
\newcommand*{\baselineB}{{Extreme-RL}\xspace}

\begin{document}

\twocolumn[{%
\renewcommand\twocolumn[1][]{#1}%
\maketitle
\vspace{-0.15in}
\begin{center}
    Project Page: \href{https://gentle-humanoid.axell.top/}{gentle-humanoid.axell.top}
\end{center}
\begin{center}
    \vspace{0.1in}
    \centering
    \includegraphics[width=\linewidth]{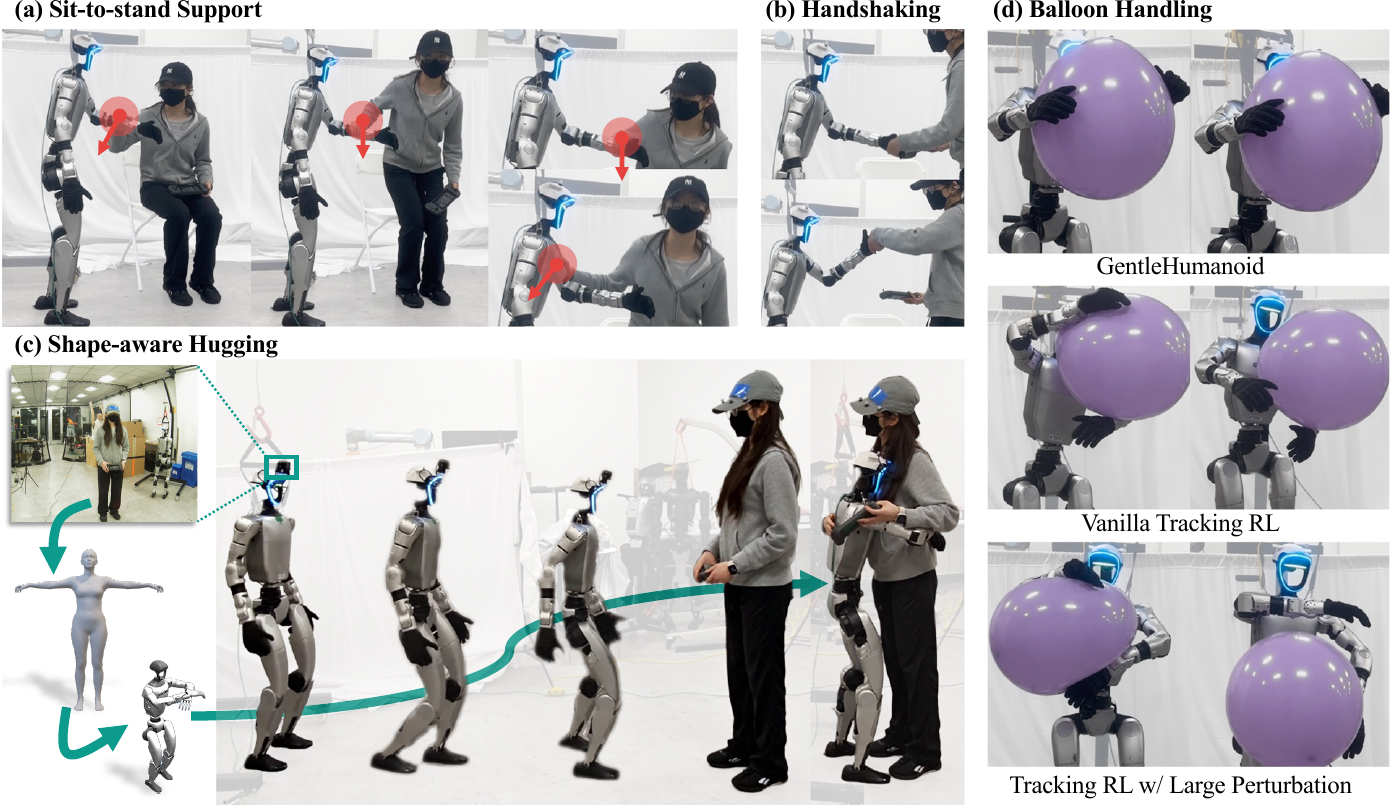}
    \captionof{figure}{\modelnametext learns a universal whole-body control policy with upper-body compliance and tunable force limits. It enables: (a) sit-to-stand assistance, where the robot provides support across multiple links (hand, elbow, and shoulder); (b) handshaking with a 5 N force limit, allowing the robot’s hand to move naturally with the human’s; (c) autonomous shape-aware hugging, where the robot adapts its posture to the partner’s body shape (estimated from camera input) for a comfortable embrace; and (d) balloon handling, showing safe object manipulation where baselines fail.}
    \label{fig:teaser}
    \vspace{0.05in}
\end{center}
}]

\thispagestyle{empty}
\pagestyle{empty}

\begin{abstract}
\renewcommand\thefootnote{}
\footnotetext{$^*$ Equal contribution. This work was done during Qingzhou Lu's internship at Stanford University. Qingzhou is now with Tsinghua University.} 
Humanoid robots are expected to operate in human-centered environments where safe and natural physical interaction is essential. However, most recent reinforcement learning (RL) policies emphasize rigid tracking and suppress external forces. Existing impedance-augmented approaches are typically restricted to base or end-effector control and focus on resisting extreme forces rather than enabling compliance. 
We introduce GentleHumanoid, a framework that integrates impedance control into a whole-body motion tracking policy to achieve upper-body compliance. At its core is a unified spring-based formulation that models both resistive contacts (restoring forces when pressing against surfaces) and guiding contacts (pushes or pulls sampled from human motion data). This formulation ensures kinematically consistent forces across the shoulder, elbow, and wrist, while exposing the policy to diverse interaction scenarios. Safety is further supported through task-adjustable force thresholds.
We evaluate our approach in both simulation and on the Unitree G1 humanoid across tasks requiring different levels of compliance, including gentle hugging, sit-to-stand assistance, and safe object manipulation. Compared to baselines, our policy consistently reduces peak contact forces while maintaining task success, resulting in smoother and more natural interactions. These results highlight a step toward humanoid robots that can safely and effectively collaborate with humans and handle objects in real-world environments. 

\end{abstract}

\section{INTRODUCTION}
Safe and compliant physical interaction is essential for deploying humanoids in human-centered environments.  
Reinforcement learning (RL) has recently enabled impressive whole-body locomotion and manipulation~\cite{human-humanoid,fu2024humanplus,ji2024exbody2,chen2025gmt,he2025asap,liao2025beyondmimic,he2024omnih2o,ze2025twist}.
However, most policies emphasize rigid position or velocity tracking and treat external forces as disturbances to suppress, which limits their applicability to tasks requiring adaptive compliance, such as handling objects.
To address this, recent works have integrated impedance or admittance control into RL~\cite{portela2024learning,facet,unifp} or attempted to learn forceful loco-manipulation implicitly~\cite{falcon}. However, these approaches are  restricted to base or end-effector control and typically emphasize resisting extreme forces rather than supporting compliant interaction. 

In contrast, interactions such as giving a comforting hug or assisting with sit-to-stand support require compliance across the entire upper-body kinematic chain, where multiple links including shoulders, elbows, and hands may be in contact simultaneously. 
Depending on the scenario, compliance must range from gentle yielding (e.g., hugging people or handling fragile objects) to firm, supportive assistance (e.g.,  sit-to-stand), while always remaining within safe force thresholds. 
This raises two main challenges: (1) coordinating force responses across multiple links of the kinematic chain, and (2) adapting to diverse contact scenarios, from gentle touch to strong supportive forces.

We address these challenges with \modelnametext, a framework that integrates impedance control into a motion-tracking policy to achieve whole-body humanoid control with upper-body compliance. The humanoid’s action is influenced by two forces: a \textit{driving force} for motion tracking, modeled as a virtual spring–damper system that pulls link positions toward target motions, and an \textit{interaction force} that represents physical contact with humans or objects. 

Since collecting real interaction data is difficult, we simulate interaction forces during RL training. Physics engines such as MuJoCo and IsaacGym can generate contact forces at colliding surfaces, but these are often noisy, local, and uncoordinated, unlike the smooth multi-joint compliance observed in human–human interactions. They also only occur when collisions arise during rollout, limiting coverage of diverse interaction scenarios.  
To address this, we introduce a unified spring-based formulation with two cases: (i) \textit{resistive contact}, when the humanoid presses against a surface, modeled by fixing the spring anchor at the initial contact point to generate restoring forces; and (ii) \textit{guiding contact}, when the humanoid is pushed or pulled by external agents, modeled by sampling spring anchors from upper-body postures in human motion datasets. Importantly, sampling from complete postures ensures forces remain coordinated across the kinematic chain (e.g., shoulder, elbow, wrist), rather than being applied independently to each link.  
This method provides kinematically consistent and diverse interaction forces, enabling the policy to learn robust compliance. 
To further ensure safety, we apply force-thresholding during training, with adjustable limits at deployment based on task requirements.

We evaluate \modelnametext against baselines, including a vanilla whole-body RL tracking policy and an end-effector-based force-adaptive policy, in both simulation and on the Unitree G1 humanoid. Quantitative tests use commercial force gauges and conformable, customized waist-mounted pressure sensing pads with 40 calibrated capacitive taxels to measure contact forces and pressures. Qualitative demonstrations cover scenarios requiring different levels of compliance, including gentle hugging, sit-to-stand assistance, and soft-object manipulation. We also show an autonomous hugging pipeline that integrates our policy with vision-based human shape estimation for personalized hugs.  

In summary, the main contributions of this work are:  
\begin{itemize}
    \item We propose \modelnametext, a framework that integrates impedance control with motion tracking to achieve whole-body humanoid control with upper-body compliance. Central to the framework is a unified formulation of interaction force modeling that covers both resistive and guiding contacts, sampling from human motion datasets to ensure kinematic consistency and capture diverse interaction scenarios.  
    \item We develop a force-thresholding mechanism that maintains interaction forces within safe limits, enabling comfortable and safer physical human–robot interaction.  
    \item We design a hugging evaluation setup with a custom pressure-sensing pad tailored for hugging, providing reliable measurement of distributed contact forces. We validate our approach in both simulation and on the Unitree G1 humanoid, showing safer, smoother, and more adaptable performance than baselines across hugging, sit-to-stand assistance, and object manipulation.  
\end{itemize}

\section{Related Work}

\subsection{Humanoid Whole Body Control}
Whole-body control for humanoid robots is a long-standing challenge in robotics. The difficulty is precipitated by high-dimensional dynamics and human-like morphology that introduces inherent instability. Traditional model-based methods, such as model predictive control (MPC), can produce stable behaviors but demand extensive expert design and meticulous tuning to balance feasibility and computational cost~\cite{murooka2021humanoid,dantec2021whole,sombolestan2024adaptive}.
More recently, learning-based methods have alleviated many of the challenges of tedious design in model-based methods. In particular, learning from human motion data has been successful for producing highly dynamic motions with single-skill policies~\cite{he2025asap} and generalist policies~\cite{ji2024exbody2,chen2025gmt,liao2025beyondmimic}. Similar frameworks have also been used for whole-body tele-operation~\cite{he2024omnih2o,fu2024humanplus,ze2025twist}.
However, these approaches often neglect scenarios involving complex contact dynamics, which reduces their robustness to external disturbances and raises safety concerns in close physical interaction with humans.

\subsection{Force-adaptive Control}
To address the aforementioned issue of robust and safe contact, classical force-adaptive methods such as impedance and admittance control regulate interaction forces and have been extended to whole-body frameworks~\cite{sombolestan2023hierarchical,sombolestan2024adaptive,rigo2024hierarchical}. More recently, RL-based approaches have incorporated impedance or admittance control for adaptive contact behaviors~\cite{portela2024learning,facet,unifp}, while others aim to implicitly learn robustness to external disturbances and extreme forces~\cite{falcon,fey2025bridging}. However, these methods typically focus on end-effector interactions rather than interactions that involve other body parts. 
In tasks such as carrying large objects or interacting with a human, contact is not restricted to the wrists/hands but may involve coordinated force distribution across multiple links, including elbows, and shoulders. Our work addresses this gap by introducing a framework that models compliance across the whole upper body kinematic chain.

\subsection{Human-humanoid Interaction}
As humanoid robots move closer to deployment in human-centered environments, their ability to interact physically with people becomes increasingly important. 
Towards this goal, early works have explored using human-in-the-loop strategies and haptic feedback to deliver soft and comfortable contact~\cite{mukai2010nursing_care,block2021huggiebot}. More recent efforts have applied traditional control methods to assist humans in specific tasks such as sit-to-stand transitions~\cite{bolotnikova2021,lefevre2024sit2stand}. However, these approaches are typically tailored to a single scenario, and the resulting policies do not generalize across different interaction contexts such as both hugging and sit-to-stand assistance. Other recent works shift the focus to vision-based criteria, for example, designing policies that enable humanoids to consistently avoid human collisions~\cite{sun2025spark}. In contrast, our approach proposes a general motion-tracking policy capable of handling multiple interaction scenarios. In particular, for hugging tasks, we combine the policy with visual perception to customize hugging positions for people of different body shapes. 

\section{Method}
\begin{figure*}[t]
    \centering
    \vspace{0.12in}
    \includegraphics[width=0.9\textwidth]{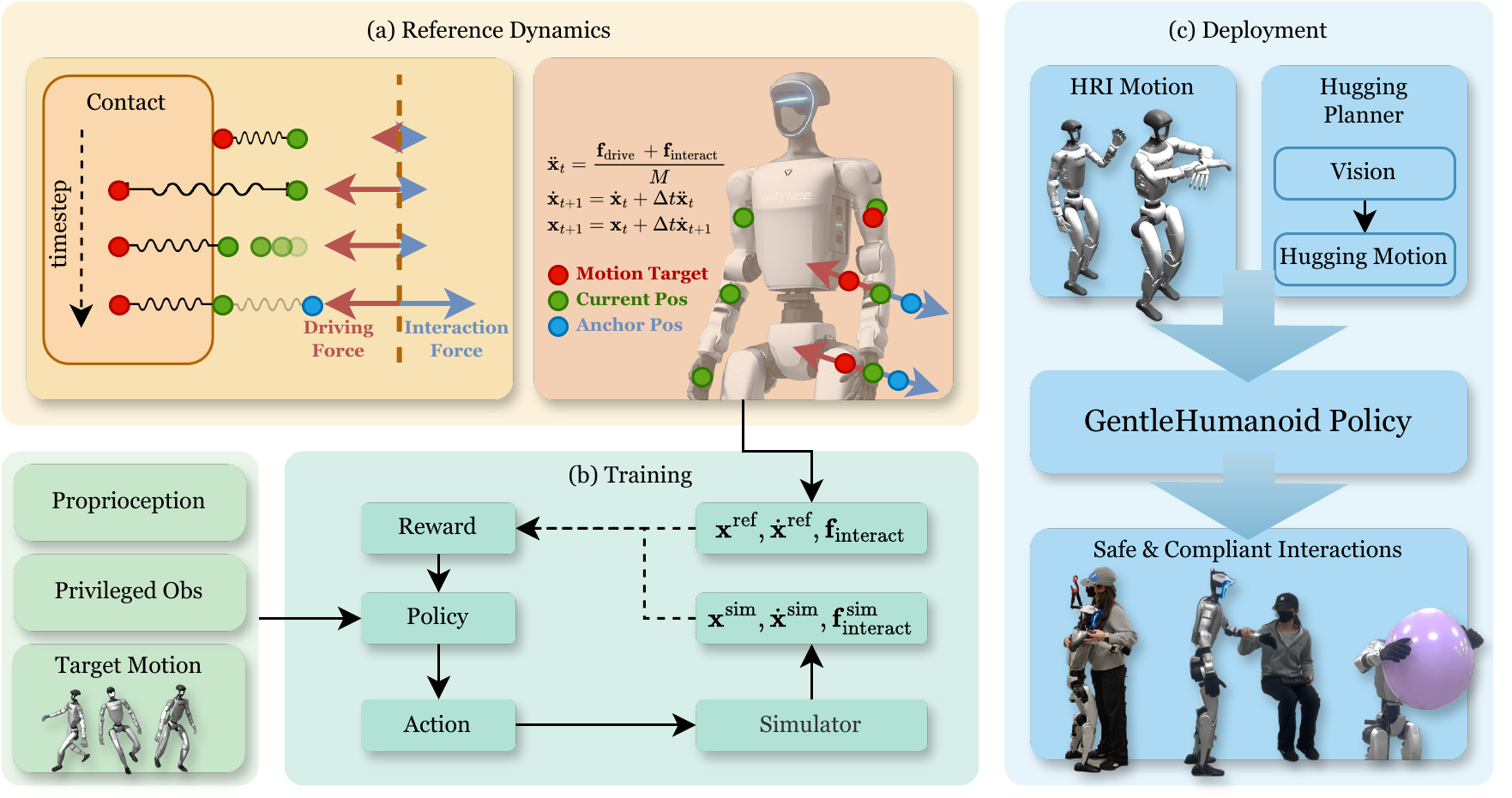}
    \caption{Overview framework. (a) Reference dynamics: impedance-based dynamics integrate driving forces (for motion tracking) and interaction forces (for compliant contact), producing reference link (on the shoulders, elbows and hands) positions and velocities. (b) Training: the policy receives proprioception, privileged observations, and target motions, and is optimized using rewards that compare simulated states $(\bm{x}^{\text{sim}}, \dot{\bm{x}}^{\text{sim}})$ to reference dynamics $(\bm{x}^{\text{ref}}, \dot{\bm{x}}^{\text{ref}})$. (c) Deployment: the trained GentleHumanoid policy is applied to real-world tasks, including vision-based autonomous hugging and other human–robot interaction scenarios, enabling safe and compliant behaviors such as hugging, sit-to-stand assistance, and handling large deformable objects.}
    \label{fig:impedance_model}
\end{figure*}

\subsection{Problem Formulation}  

Our goal is to achieve whole-body humanoid control that is both robust and safe, enabling humanoids to perform diverse motions while interacting compliantly with humans and deformable objects. We frame this as learning a compliant motion-tracking policy: the humanoid should follow human-like movements while adapting its behavior in response to interaction forces. Unlike rigid trajectory tracking, humans naturally adjust their actions based on contact feedback, which motivates our use of impedance-based control.  
Since most physical interactions occur in the upper body, we focus on modeling it as a multi-link impedance system with keypoints at the shoulders, elbows, and hands. 
As illustrated in Fig.~\ref{fig:impedance_model}, the motion of each link position is influenced by the combination of driving forces from target motions and interaction forces from humans or objects:  
\begin{equation}
M\ddot{\bm{x}}_i = \bm{f}_{\text{drive},i} + \bm{f}_{\text{interact},i} \;,
\end{equation}
where $\bm{x}_i$ is the position of link $i$, $\ddot{\bm{x}}_i$ is acceleration, and $M$ is a scalar virtual mass (kg) per link.  
We set $M$ as $0.1$ kg in our reference dynamics model. The driving force $\bm{f}_{\text{drive},i}$ is a virtual spring–damper term from classical impedance control, pulling the link position toward its target motion, and $\bm{f}_{\text{interact},i}$ captures forces arising from interactions with the environment, including humans and objects. 
In the following sections, we detail the formulation of each force component. For clarity, we introduce the index $i$ once and omit it henceforth. 
All link positions $\bm{x}$ and velocities $\dot{\bm{x}}$ are 3D Cartesian quantities expressed in the robot's root frame.

\subsection{Impedance-Based Driving Force from Target Motion}  

Following prior work~\cite{sombolestan2024adaptive,facet}, we generate driving forces from the target motion to pull each link position toward its target trajectory. The force is modeled as a virtual spring–damper system:  
\begin{equation}
\bm{f}_{\text{drive}} = K_p(\bm{x}_{\text{tar}} - \bm{x}_{\text{cur}}) + K_d(\bm{v}_{\text{tar}} - \bm{v}_{\text{cur}})\;,
\label{eq:driving_force}
\end{equation}
where $\bm{x}_{\text{cur}}, \bm{v}_{\text{cur}}$ are the current link position and velocity, and $\bm{x}_{\text{tar}}, \bm{v}_{\text{tar}}$ are the corresponding target link position and velocity from the target motion. The gains $K_p$ and $K_d$ denote the impedance stiffness and damping, respectively, controlling how strongly the link position tracks its target. 
To ensure stable and smooth behavior, we set the damping to the critical value, $K_d = 2\sqrt{M K_p}$.  
All $\bm{x}$ and $\bm{v}$ terms above denote 3D Cartesian link states (in the root frame), while the policy produces actions in joint space that are tracked by low-level joint PD controllers. The RL policy learns to coordinate these compliant forces across multiple joints, mapping them into joint-level actions that balance stability and adaptability in whole-body control. 

\subsection{Interaction Force Modeling}  

When no interaction occurs, the driving force alone enables the humanoid to follow target motions. In real scenarios, however, physical contact introduces additional interaction forces across multiple links, often correlated in direction and magnitude. To capture these effects, we design a unified interaction force model that accounts for both multi-link coupling and force diversities. We distinguish two cases:  

\textbf{Resistive contact}: Forces generated when the humanoid itself presses against a human or object. 

\textbf{Guiding contact}: Forces applied by an external agent, such as a human pushing or pulling the humanoid’s arm.  

Both cases are modeled using the same spring formulation with a consistent anchor terminology:  
\begin{equation}
    \bm{f}_{\text{interact}} = K_{\text{spring}} \bigl(\bm{x}_{\text{anchor}} - \bm{x}_{\text{cur}}\bigr),
\end{equation}
where $K_{\text{spring}}$ is the stiffness, $\bm{x}_{\text{cur}}$ is the current link position, and the spring anchor $\bm{x}_{\text{anchor}}$ is defined as
\begin{equation}
    \bm{x}_{\text{anchor}} =
    \begin{cases} 
        \bm{x}_{\text{cur}}(t_{0}), & \text{resistive contact}, \\[6pt]
        \bm{x}_{\text{sample}}, & \text{guiding contact}.
    \end{cases}
\end{equation}
Here, $\bm{x}_{\text{cur}}(t_{0})$ is the link position at the moment of initial contact (fixing a virtual spring anchor), $\bm{x}_{\text{sample}}$ is a link position sampled from a dataset posture, representing an external agent steering the humanoid toward a new configuration. 

This formulation provides a unified framework: \textbf{Resistive contact} yields restoring forces that resist deviations from the contact point, while \textbf{Guiding contact} yields guiding forces that pull the humanoid toward externally defined postures. 
Posture samples are drawn from real human motion data, ensuring that the guiding forces are kinematically valid and correspond to plausible upper-body movements. Specifically, we precompute posture distributions from motion dataset, during training, select postures close to the current multi-link positions. From these, a target position is randomly sampled and used as the spring anchor to generate guiding forces. 

\begin{figure}[t]
    \centering
    \includegraphics[width=1.0\linewidth]{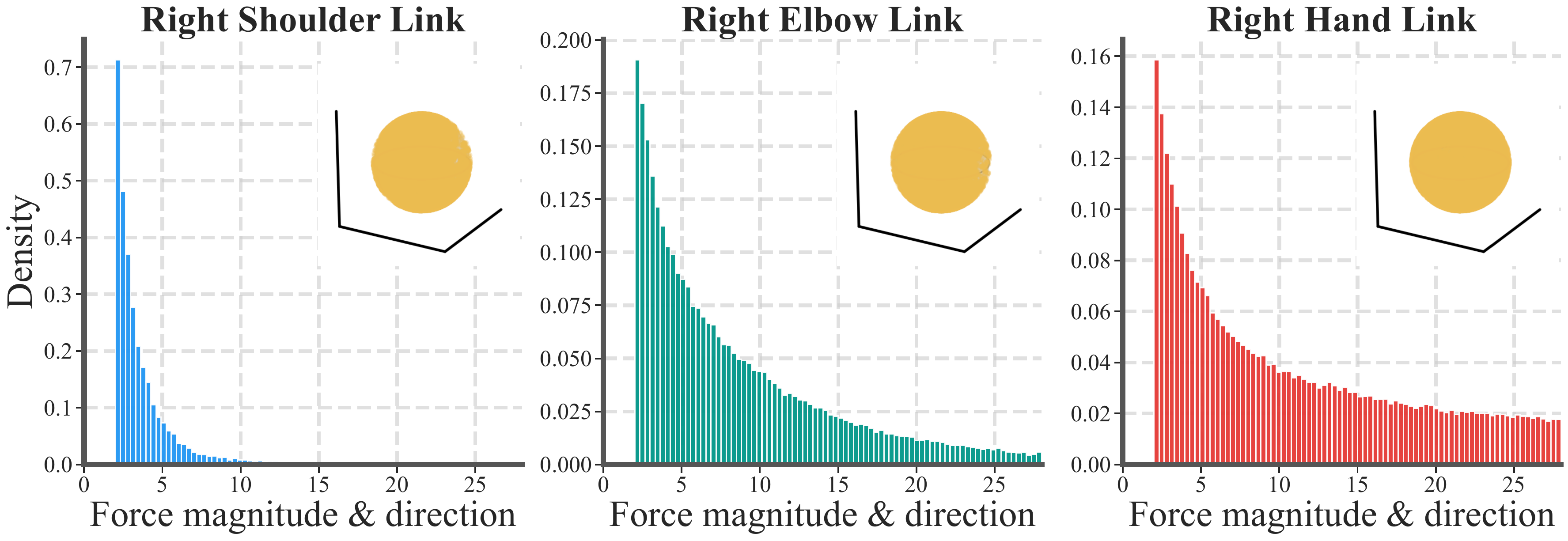}
    \caption{Interaction force distributions across upper-body links. Probability densities of force magnitudes are shown for the right shoulder (left), right elbow (middle), and right hand (right). Insets (top right) illustrate the corresponding force directions on a sphere.}
    \label{fig:method_force_distribution}
\end{figure}

To further increase interaction diversity, we randomize both stiffness and the active links. The stiffness is sampled as $K_{\text{spring}} \sim \mathcal{U}(5,\,250)$. 
Active-contact sets are chosen with the following probabilities: 40\% no external force; 15\% both arms (all 6 links) under force; 30\% a single arm (left or right; its 3 links) under force (15\% each arm); and 15\% only a single link under force. Anchors and selections are resampled every 5 seconds with a short transition window to ensure continuity. This exposes the policy to a broad range of interaction dynamics, enabling it to learn robust compliance while preserving consistency along the kinematic chain.
As a result, the model can simulate diverse external force directions and magnitudes; Figure~\ref{fig:method_force_distribution} visualizes the resulting distribution, showing that forces span a wide range of directions on the sphere with magnitudes from 0 to 25 N. 

\subsection{Safety-Aware Force Thresholding}  

In Equation~\ref{eq:driving_force}, the driving force grows proportionally with tracking error. Without limitation, large deviations from the target motion can result in unbounded forces, potentially exceeding safe interaction levels. To prevent this, we introduce an adaptive force thresholding mechanism that caps the maximum allowable force applied by the robot.  

We define a range of force thresholds and sample a piecewise-constant value $\tau_{\text{safe}}$ during training: $F_1 \leq \tau_{\text{safe}} \leq F_2$. The threshold is resampled every 5 seconds, encouraging the policy to remain robust across a range of safety limits. The current threshold is also provided to the policy as part of the observation.
Here, $F_1$ and $F_2$ define the range for the maximal allowable force the robot should apply in various tasks. 
When the driving force exceeds the threshold, we apply a scaling mechanism:  
\begin{equation}
\bm{f}_{\text{drive\_limited}} = \min\left(1.0, \frac{\tau_{\text{safe}}}{\|\bm{f}_{\text{drive}}\|}\right) \cdot \bm{f}_{\text{drive}}\;,
\label{eq:safe_thresholding}
\end{equation}
compliance. The threshold directly tunes compliance: lower values yield softer, safer behavior for gentle interactions like hugging, while higher values allow firmer support for tasks such as sit-to-stand assistance, all while maintaining safety bounds. 
The choice of exact threshold depends on the application. Since our focus is humanoid interaction with humans and fragile objects (e.g., balloons), we set $F_1=5$~N and $F_2=15$~N. These values are benchmarked against both ISO/TS~15066~\cite{ISO/TS15066:2016} safety ceilings and comfort studies. In the extreme case of a minimal $0.5 \times 0.5$~cm contact area ($0.25~\text{cm}^2$), $15$~N corresponds to $60~\text{N/cm}^2$, still below ISO/TS~15066 pain-onset limits for torso and arms (e.g., back/shoulder: $160~\text{N/cm}^2$, chest: $120~\text{N/cm}^2$). For more realistic hugging contacts of $\sim 16~\text{cm}^2$, this range corresponds to $3$--$9$~kPa, consistent with measurements of children’s hugs (soft hugs $<7$~kPa, strong hugs $\approx 18$~kPa)~\cite{disney_hug} and rehabilitation studies recommending pressures $\leq 13$~kPa for comfort~\cite{rehab_pressure}.   
Thus, our thresholds remain well below ISO ceilings while lying in a comfort-oriented band.  

\subsection{RL-based Control Policy} 

Formally, we consider a humanoid robot at time $t$ with observation $\bm{o}_t$ containing its proprioception and a target motion sequence $\bm{m}_{tar}$.
The policy $\pi(\bm{a}_t \mid \bm{o}_t)$ outputs joint position targets $\bm{a}_t$ at $50$ Hz for low-level PD tracking, enabling the humanoid to follow the target motion while exhibiting compliant responses to interaction forces $\bm{f}_{\text{interact}}$.  

To incorporate the impedance-based reference dynamics, we simulate the model using semi-implicit Euler integration, with a fixed time step of $0.005$ s:  
\begin{align}
\dot{\bm{x}}_{t+1}^{\text{ref}} &= \dot{\bm{x}}_{t}^{\text{ref}}
+ \Delta t \cdot \tfrac{\bm{f}_{\text{drive}} + \bm{f}_{\text{interact}}}{M}, \\
\bm{x}_{t+1}^{\text{ref}} &= \bm{x}_{t}^{\text{ref}}
+ \Delta t \cdot \dot{\bm{x}}_{t+1}^{\text{ref}}.
\label{eq: reference_dyanmic_model}
\end{align}
Where $\Delta t$ is the integration step size, and $\bm{x}_{t}^{\text{ref}}$ denotes the link position in the reference dynamics model, which we distinguish from the actual robot link position $\bm{x}^{\text{sim}}$ in the simulator. The objective is to guide the robot to follow the impedance rules encoded in the reference dynamics. 
At each timestep, velocities and positions are updated according to the net driving and interaction forces, with semi-implicit Euler ensuring numerical stability.  

This impedance-based reference dynamics system specifies the compliant behavior the policy is trained to reproduce. We compute $\bm{x}^{\text{ref}}$ via the above integration and use it in the link-position tracking rewards (details in Reward Design). During training, the RL agent observes $\bm{o}_t$ and outputs $\bm{a}_t$ such that the resulting behavior aligns with this dynamics model. In effect, the policy learns to track target motions while adapting to stochastic interaction forces, yielding stable, compliant whole-body control across diverse scenarios. 

\subsubsection{Teacher-Student Architecture}

We employ a two-stage teacher–student training framework for sim-to-real transfer. We adopt the same teacher-student architecture and training procedure from prior work~\cite{facet}, and train both policies with PPO~\cite{PPO}. The student policy observes only information available during real-world deployment: 
$$\bm{o}_t = (\tau_{\text{safe}}, \bm{m}_{\text{tar}}, \boldsymbol{\omega}, \bm{g}, \bm{q}_{t}^{\text{hist}}, \bm{a}_{t-3:t-1})\;,$$
where $\tau_{\text{safe}}$ represents the current force-safety limit, that can be changed by use during deployment;  
$\bm{m}_{\text{tar}}$ contains target motion information including future root poses and target joint position; 
$\boldsymbol{\omega}$ is the root angular velocity; and $\bm{g}$ is gravity expressed in the robot's root frame (projected gravity). $\bm{q}_{t}^{\text{hist}}$ provide joint-position history, and $\bm{a}_{t-3:t-1}$ contains the recent action history. 

The teacher policy additionally receives comprehensive privileged information:
$$
\bm{o}_t^{\text{priv}} = (\bm{x}^{\text{ref}}_t, \dot{\bm{x}}^{\text{ref}}_t, 
\bm{f}_{\text{interact}}, \bm{f}_{\text{interact}}^{\text{sim}}, 
\bm{h}_t, \boldsymbol{\tau}_{t-1}, \bm{e}_{\text{cum}})\;,
$$ 
where $\bm{x}^{\text{ref}}_t$ and $\dot{\bm{x}}^{\text{ref}}_t$ are the integrated link positions and velocities from the impedance-based reference dynamics (Eq.~\ref{eq: reference_dyanmic_model});  
$\bm{f}_{\text{interact}}$ denotes the interaction force predicted by the reference dynamics, while $\bm{f}_{\text{interact}}^{\text{sim}}$ is the actual interaction force measured in simulation. Ideally, $\bm{f}_{\text{interact}}$ should closely match $\bm{f}_{\text{interact}}^{\text{sim}}$.  
$\bm{h}_t$ represents link heights relative to the ground; $\boldsymbol{\tau}_{t-1}$ are the previous joint torques; and $\bm{e}_{\text{cum}}$ denotes the cumulative tracking error.  

Both policies output joint position targets $\bm{a}_t \in \mathbb{R}^{29}$ which are tracked by low-level PD controllers.

\subsubsection{Motion Datasets}

We use diverse human motion to train our policy, covering data for both human-human and human-object interactions datasets. Specifically, we use GMR~\cite{ze2025gmr} to retarget the AMASS~\cite{AMASS}, InterX~\cite{xu2023interx}, and LAFAN~\cite{harvey2020robust} datasets, and filter out some high-dynamic motions that do not conform to interaction scenarios, ultimately obtaining approximately 25 hours of dataset with a sampling frequency of 50Hz. 

\subsubsection{Reward Design}  
Following prior work on whole-body humanoid control~\cite{ze2025twist,fu2024humanplus}, we adapt rewards for motion tracking and locomotion stability, as summarized in Table~\ref{tab:rewards_unified}, to encourage accurate motion tracking and stable balance.  
In \modelnametext, we additionally design a compliance reward composed of three terms:  

\textbf{Reference Dynamics Tracking.}  
We encourage the robot to follow the compliant reference dynamics by minimizing the discrepancy between the actual link state in simulation $(\bm{x}_t^{\text{sim}}, \dot{\bm{x}}_t^{\text{sim}})$ and the reference state $(\bm{x}_t^{\text{ref}}, \dot{\bm{x}}_t^{\text{ref}})$ from Eq.~\ref{eq: reference_dyanmic_model}:   
\[
r_{\text{dyn}} = \exp\!\left(-\frac{\|\bm{x}_t^{\text{sim}} - \bm{x}_t^{\text{ref}}\|_2}{\sigma_x}\right) 
+ \exp\!\left(-\frac{\|\dot{\bm{x}}_t^{\text{sim}} - \dot{\bm{x}}_t^{\text{ref}}\|_2}{\sigma_v}\right).
\]  

Exponential kernels provide smooth gradients, with $\sigma_x$ and $\sigma_v$ controlling sensitivity.

\textbf{Reference Force Tracking.}  
To align predicted interaction forces with actual forces measured in simulation, we penalize the discrepancy between $\bm{f}_{\text{interact}}$ from the reference dynamics and $\bm{f}_{\text{interact}}^{\text{sim}}$ from the environment:   
\[
r_{\text{force}} = \exp\!\left(-\frac{\|\bm{f}_{\text{interact}} - \bm{f}_{\text{interact}}^{\text{sim}}\|_2}{\sigma_f}\right).
\]  

This term complements position tracking by explicitly regulating force magnitudes, which is crucial for enforcing safe maximum force thresholds.  

\textbf{Unsafe Force Penalty.}  
To further discourage unsafe behaviors, we penalize interaction forces that exceed the safety margin $\tau_{\text{safe}}$, in addition to the driving force thresholding in Eq.~\ref{eq:safe_thresholding}:   
\[
r_{\text{pen}} = - \mathbb{I}\!\left(\|\bm{f}_{\text{interact}}\| > \tau_{\text{safe}} + \delta_{\text{tol}}\right).
\]  

Here, $\delta_{\text{tol}}$ is a tolerance margin that allows minor deviations beyond $\tau_{\text{safe}}$ without triggering large penalties. This prevents the policy from becoming overly conservative while still discouraging forces that are clearly unsafe. In practice, we set $\delta_{\text{tol}}$ as $10$~N based on empirical observations.  

The overall compliance reward is a weighted sum of these terms:  
\[
r_{\text{compliance}} = w_{\text{dyn}} \, r_{\text{dyn}} 
+ w_{\text{force}} \, r_{\text{force}} 
+ w_{\text{pen}} \, r_{\text{pen}} .
\]  
The weights for each term along with those for motion tracking and locomotion stability are provided in Table~\ref{tab:rewards_unified}.

\begin{table}[h!]
\caption{Reward Terms and Weights.}
\label{tab:rewards_unified}
\centering
\begin{tabular}{lc}
\toprule
\textbf{Reward} & \textbf{Weight} \\
\midrule
\multicolumn{2}{l}{\textbf{Compliance}} \\
\midrule
Reference Dynamics Tracking     & 2.0 \\
Reference Force Tracking              & 2.0 \\
Unsafe Force Penalty             & 6.0 \\
\midrule
\multicolumn{2}{l}{\textbf{Motion Tracking}} \\
\midrule
Root Tracking             & 0.5 \\
Joint Tracking            & 1.0 \\
\midrule
\multicolumn{2}{l}{\textbf{Locomotion Stability}} \\
\midrule
Survival                  & 5.0 \\
Feet Air Time             & 10.0 \\
Impact Force              & 4.0 \\
Slip Penalty            & 2.0 \\
Action Rate      & 0.1 \\
Joint Velocity      & 5.0e-4 \\
Joint Limit           & 1.0   \\
\bottomrule
\end{tabular}
\end{table}


\section{Experiments}
We conduct both simulation and real-world experiments to evaluate the effectiveness of \modelname. We compare against two baselines that adopt different training strategies: 
\textbf{\baselineA}: an RL-based motion tracking policy trained without force perturbations, representative of prior whole-body tracking approaches; 
\textbf{\baselineB}: an RL-based motion tracking policy trained with maximum $30$~N end-effector force perturbations, representative of prior force-adaptive methods.

\subsection{Simulation Results}  
\begin{figure}[t]
    \centering
    \vspace{0.12in}
    \includegraphics[width=1.0\linewidth]{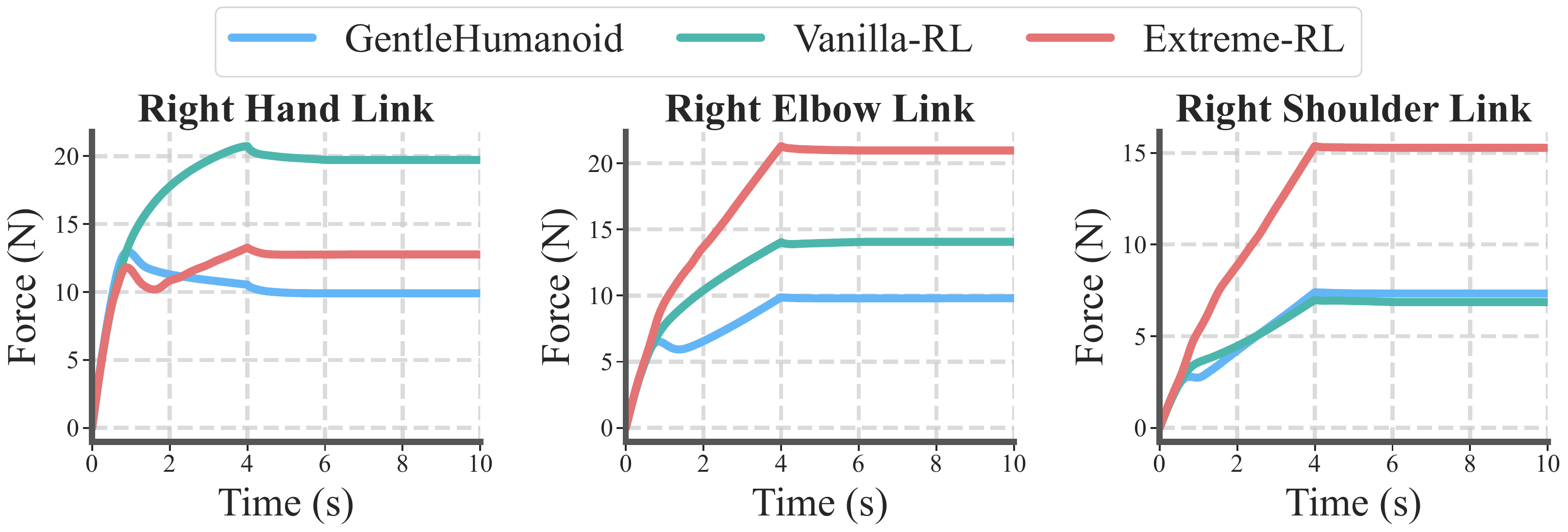}
    \caption{Forces applied by different upper-body links under external interaction. Force profiles over time are shown for the right hand (left), right elbow (middle), and right shoulder (right). Compared to baselines (\baselineA and \baselineB), \modelname maintains lower and more stable force levels across all links, showing safer and more compliant responses during contact.}
    \label{fig:sim}
\end{figure}

We first benchmark against baselines in simulation using a hugging motion. To evaluate compliance, we simulate an external pulling force that attempts to move the robot away from its hugging posture, mimicking a human trying to break free from an embrace. As shown in Figure~\ref{fig:sim}, our method consistently maintains lower and more stable interaction forces across the hand, elbow, and shoulder links. At the hand, \modelname stabilizes around 10 N, whereas \baselineA settles above 20 N and \baselineB exceeds 13 N. Similar trends are observed at the elbow and shoulder: while baselines quickly saturate at 15–20 N with rigid responses, \modelname remains bounded near 7–10 N. These results show that our method adapts smoothly to external interaction, yielding compliant motions, while baselines remain overly stiff and exert higher peak forces. 

\subsection{Real-World Experiments} 
\begin{figure}[t]
  \centering
  \includegraphics[width=\linewidth]{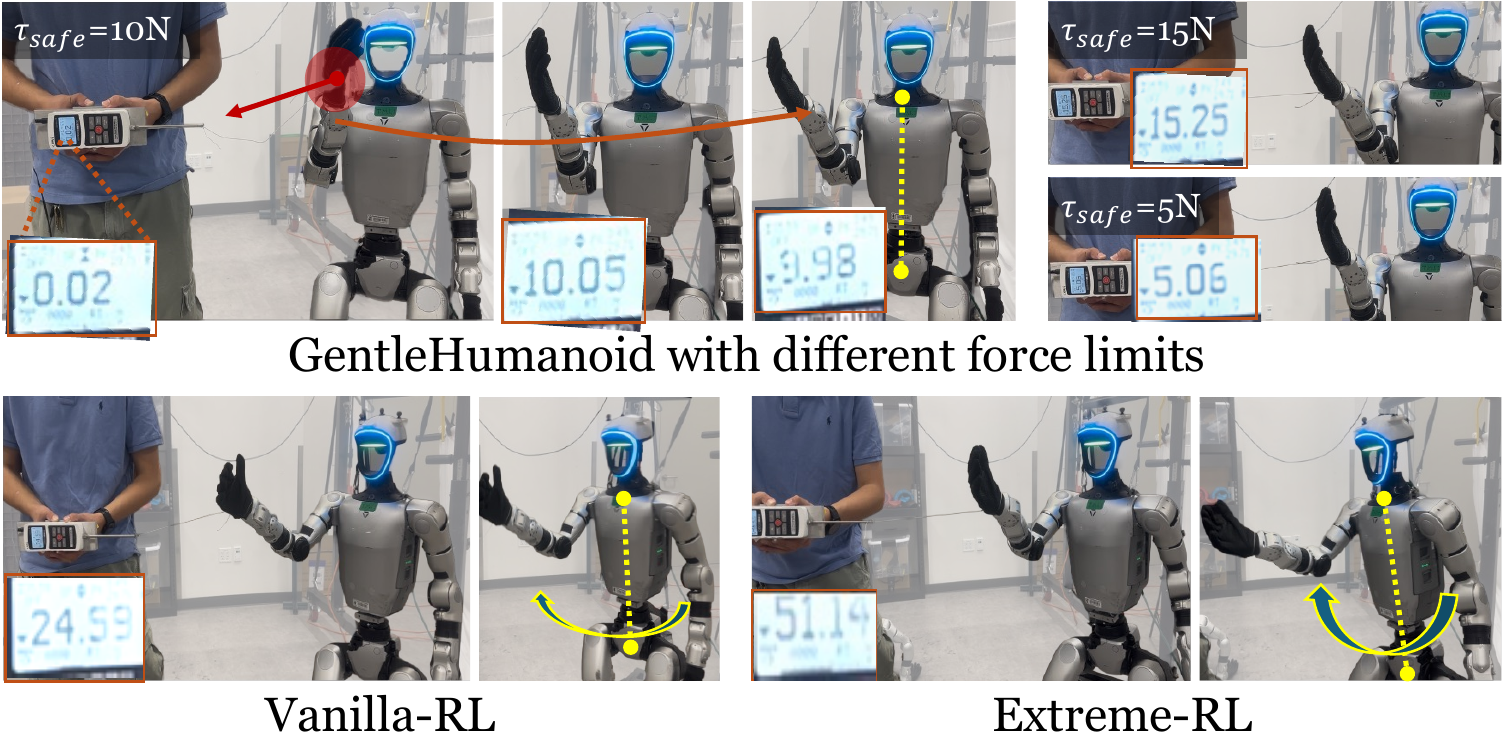}
  \caption{
  Comparison of interaction forces across policies. Top: \modelname with tunable force limits, which maintains safe interaction by keeping contact forces within specified thresholds across different postures. Bottom: baseline methods, \baselineA and \baselineB, exhibit less consistent compliance, with higher peak forces or oscillatory responses. Force gauge readings (N) are highlighted for clarity. 
  }
  \label{fig:exp_force_gauge}
\end{figure}

We deploy our whole-body control policy on the Unitree G1 humanoid to evaluate compliance in real-world interactions. Three reference scenarios are considered:  

\textbf{1) Static pose with external force.}  
We first test compliance by applying external forces at the wrist while the robot's base remains static. Ideally, the arm should yield softly, moving with the external force instead of resisting rigidly. Forces are applied using a handheld force gauge (Mark-10, M5-10), which also records peak values.  
As shown in Figure~\ref{fig:exp_force_gauge}, both baselines resist stiffly: rather than letting the arm move, the torso shifts, often leading to imbalance. \baselineB is particularly rigid, requiring a peak force of 51.14~N, while \baselineA requires 24.59~N. In contrast, \modelname responds smoothly and consistently, requiring much lower forces to reposition the arm while maintaining balance.  
A key observation is that \modelname provides posture-invariant compliance: the same external force suffices to modulate arm position across different configurations. Moreover, compliance level matches the user-specified force limit. For example, when set to 10~N, the robot maintains balance around that threshold across postures, with effective ranges between 5–15~N. This uniform, predictable response arises from our formulation, which regulates compliance through virtual spring–damper dynamics and safety thresholds rather than raw joint mechanics. As a result, human interaction feels safer and more consistent than with baselines.  

\begin{figure}[t]
  \centering
  \vspace{0.1in}
\includegraphics[width=0.95\linewidth]{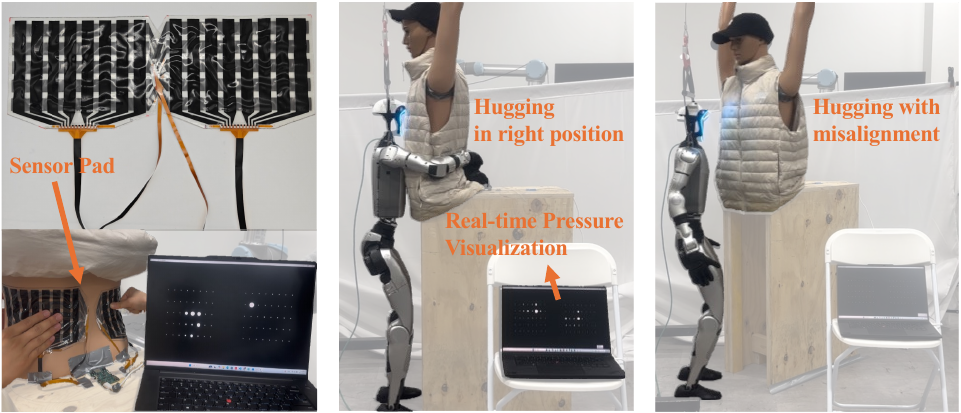}
\includegraphics[width=\linewidth]{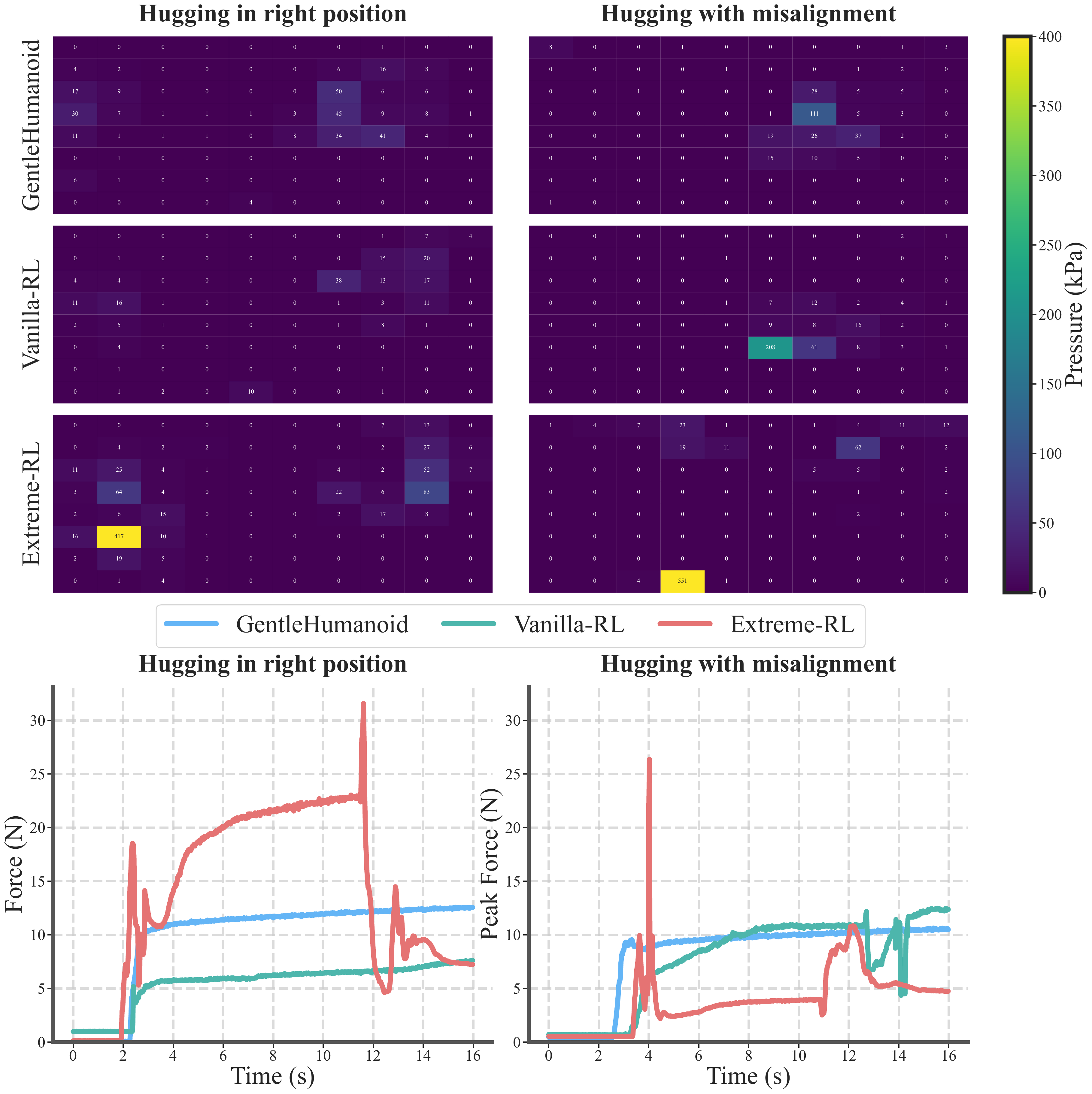}
  \caption{Evaluation of hugging interactions with and without misalignment. 
  Top: experimental setup with custom pressure-sensing pads and real-time pressure visualization. 
  Middle: pressure maps of peak force frames for different controllers  under correct hugging alignment (left) and misalignment (right). GentleHumanoid maintains moderate contact pressures, while baselines produce localized high-pressure peaks, especially under \baselineA. 
  Bottom: Force profiles over time, where \modelname maintains bounded and stable forces, while baselines exhibit increasing or unstable peaks. 
  }
  \label{fig:exp_hugging}
\end{figure}

\textbf{2) Hugging a mannequin.}  
We next evaluate hugging performance under two conditions. In the first, the mannequin is properly aligned with the robot, and the G1 executes a hugging motion. In the second, the mannequin is deliberately misaligned to assess safety under imperfect contact.
Pressure-sensing pads attached to the mannequin measure contact forces. We set $\tau_{\text{safe}}$ as 10~N in \modelnametext to compare with baselines. 
For sensor calibration, a motorized stage with a PDMS applicator was used to map normalized sensor values to ground-truth pressures measured by a force gauge. Under localized contact, we approximate the effective contact area of each texel as $6\,\text{mm} \times 6\,\text{mm}$ and compute forces from the corresponding pressure values recorded in the pad.   
The evaluation setups and results are shown in Figure~\ref{fig:exp_hugging}, \modelname maintains bounded and stable forces even under misalignment, whereas the baselines \baselineA and \baselineB generate higher, less predictable forces or fail to sustain the motion. 


\textbf{3) Handling deformable objects.}  
Finally, we test the ability to handle fragile objects such as balloons. The challenge is to maintain contact forces within a safe range: insufficient force fails to stabilize the object, while excessive force causes deformation or collapse. For this experiment, the force threshold in \modelnametext is set to 5~N. As shown in Figure~\ref{fig:teaser}(d), \modelname successfully holds the balloon without damage, whereas both baselines apply excessive pressure, eventually squeezing the balloon until the G1 loses balance and drops it.  

Across all scenarios, \modelname consistently reduced peak interaction forces compared to baselines, resulting in safer and smoother contact. 
\subsection{More Applications}  
\modelnametext enables applications where compliance is critical. We integrate our policy with a locomotion teleoperation framework for the Unitree G1, allowing users to control walking and trigger pre-defined reference motions such as hugging, sit-to-stand assistance, and object handling. Demonstrations of joystick-based control are provided in the supplementary video. While this work focuses on locomotion teleoperation, extending \modelnametext to full-body teleoperation such as TWIST~\cite{ze2025twist} is an important direction for future work. The inherent compliance of our method ensures safe interactions even during teleoperation under direct physical contact, making it particularly promising for healthcare and assistive scenarios where caregivers or operators remotely guide humanoid motions. 

We also develop an autonomous, shape-aware pipeline for personalized hugging. The human’s location and height are obtained using a motion-capture system with markers placed on a hat, while an additional RGB camera mounted on the G1’s head provides input for single-image human shape estimation, as shown in Figure~\ref{fig:teaser}(c). From this image, we reconstruct a personalized body mesh using an existing human mesh estimation method~\cite{bedlam} and scale it to the subject’s true height. Waist points are then extracted from the mesh to optimize the humanoid’s hugging motion by aligning its hands with these target locations. This allows the G1 to adapt its hugging posture to individuals of different body shapes in a fully autonomous manner. Experiments with participants of varying heights and builds show that the pipeline generates stable and comfortable hugging motions.

\section{Discussion and Limitations} 
Our study shows that \modelnametext enables upper-body compliance in humanoid robots. By integrating impedance control into whole-body motion tracking and training with a unified spring-based formulation, the policy generates coordinated responses across multiple links and reduces peak contact forces compared to baselines. Demonstrations in hugging, sit-to-stand assistance, and object handling highlight its ability to adapt compliance across diverse scenarios, underscoring its potential for human-centered interaction.

Several limitations remain. First, we use human motion data to maintain kinematic consistency across links, but the dataset itself constrains the force distribution. For instance, forces applied to the shoulder are relatively small due to limited variation in the recorded motions. Incorporating more diverse motion datasets, such as dancing, could further improve coverage.  Second, our interaction modeling relies on simulated spring forces, which provide structured coverage and kinematic consistency but do not fully capture the complexity of real human contact, such as frictional effects or the viscoelastic properties of human tissue.  
Third, although the safety-aware policy constrains interaction forces, real-world experiments reveal occasional overshoots of 1–3~N due to sim-to-real discrepancies. Additional tactile sensing may be necessary for more precise force regulation.  
Finally, human localization and height are currently obtained from a motion capture system. Replacing this with a vision-based pipeline would improve autonomy and practicality, particularly in long-horizon tasks.  
Future work will focus on integrating richer sensing, combining general perception and reasoning systems such as vision language models, and extending evaluations to long-horizon interactions where the humanoid must adapt its motion dynamically to human partners’ behaviors.

\section{Acknowledgment}
We would like to thank Haoyang Weng, Botian Xu, Haochen Shi, Sirui Chen, Ken Wang, Yanjie Ze, Joao Pedro Araujo, Yufei Ye and Takara Everest Truong for their valuable discussions. We are also grateful to Yu Sun for assistance with motion capture from video and to Jiaxin Lu for support with the motion dataset. We further thank the Unitree team for their timely and reliable hardware support.

\bibliographystyle{IEEEtran}
\bibliography{main}
\clearpage
\appendix

\subsection{External Force Application Logic}

We apply interaction forces at a subset of upper-body links (shoulders, wrists, hands). The procedure runs every simulation step and consists of: (i) selecting which links are currently active and their interaction spring gains, (ii) updating an anchor (spring origin), (iii) computing interaction forces in the robot root frame and integrating the compliant reference, and (iv) applying forces/torques in the simulator.

\subsubsection{Activation and Gain Scheduling}

An \emph{active link} is a force-application point that is enabled in the current interval; we denote the active set by a binary mask $\mathbf{m} \in \{0,1\}^{M}$ over the $M$ candidate links. At the beginning of an interval we sample one of five modes (no-force, all-links, left-only, right-only, or a random partial subset) to determine $\mathbf{m}$. For every active link we assign an interaction spring gain $K_{\!\text{spring}}(t)$ that varies \emph{smoothly over time} (piecewise-linear in discrete steps). Gains may gently increase, hold, and then decrease back to zero at the end of the interval.

In parallel, a force safety threshold $\tau_{\text{safe}}(t)$ is adjusted smoothly within a bounded range and later used for clamping and reward shaping.

\subsubsection{Anchor (Interaction Spring Origin) Update}

Each active link maintains an anchor $\bm{o}(t)$ in the robot root frame. We use two behaviors consistent with the two interaction types introduced: (1) \textbf{Resistive contact}: the anchor remains at its previously established location (relative to the root), modeling a resisting load at the current contact site; (2) \textbf{Guiding contact}: the anchor is smoothly moved toward a newly sampled surface point. In both cases the updates are smooth, avoiding discontinuities when the active set or targets change.

\subsubsection{One-Sided Projection}

We model contact as \emph{one-sided}: interaction forces only act when the link compresses toward the anchor along the intended direction of interaction; when the link moves away (i.e., leaves the contact side), the interaction force drops to zero. Practically, we compute the displacement from the link to the anchor, take only its component along the intended direction. This prevents non-physical pull-back in free space and emulates real unilateral contacts.

\subsubsection{Application in the Simulator}

Forces are applied in world coordinates at the active links. To prevent excessive overall disturbance, we bound the net wrench about the torso: we sum all per-link forces/torques, and if the totals exceed preset limits, we inject an opposite residual on the torso.

\begin{table}[h!]
\caption{External Force Application Parameters.}
\label{tab:ext-force-params}
\centering
\begin{tabular}{lll}
\toprule
\textbf{Parameter} & \textbf{Symbol} & \textbf{Typical value / range} \\
\midrule
Max per-link force cap & $F_{\max}$ & 30\,N \\
Safety threshold (per link) & $\tau_{\text{safe}}(t)$ & 5--15\,N (default 10\,N) \\
Net force limit (about torso) & $\tau_{\!F}$ & 30\,N \\
Net torque limit (about torso) & $\tau_{\!M}$ & 20\,N\,m \\
Interaction spring gain & $K_{\!\text{spring}}(t)$ & 5--250 \\
\bottomrule
\end{tabular}
\end{table}

\subsection{Reference Dynamics Integration}
\label{sec:ref-dyn-integration}

All reference quantities are expressed in the robot root frame. Let $\bm{x}_t,\dot{\bm{x}}_t$ be the current link state and $\bm{x}^{\text{tar}}_t,\dot{\bm{x}}^{\text{tar}}_t$ the target state. The reference dynamics used in this work are
\begin{equation}
M\ddot{\bm{x}}_t \;=\; \bm{f}_{\text{drive}}(\bm{x}^{\text{tar}}_t,\dot{\bm{x}}^{\text{tar}}_t,\bm{x}_t,\dot{\bm{x}}_t)\;+
\bm{f}_{\text{interact}}(\cdot)\;-
D\,\dot{\bm{x}}_t\,.
\end{equation}
The driving and interaction forces follow the definitions in the method, and $D\,\dot{\bm{x}}_t\,$ is an additional damping term for stability. We integrate this system with explicit Euler using a small fixed number of substeps per simulator step (four substeps in our implementation), and clip acceleration/velocity at each step. 

\begin{table}[h!]
\caption{Reference Dynamics and Integration Parameters.}
\label{tab:ref-dyn-params}
\centering
\begin{tabular}{lll}
\toprule
\textbf{Parameter} & \textbf{Symbol} & \textbf{Value} \\
\midrule
Virtual mass & $M$ & 0.1\,kg \\
Integration damping & $D$ & 2.0 \\
Tracking stiffness & $K_p$ & Derived from $K_p = \tau_{\text{safe}}/0.05$ \\
Tracking damping & $K_d$ & $2\sqrt{M K_p}$ \\
Time step & $\Delta t$ & Same as simulation $\,\mathrm{d}t = 0.02s$ \\
Substeps per simulator step & $N_{\text{sub}}$ & 4 \\
Velocity clip & $\|\dot{\bm{x}}\|_{\max}$ & 4\,m/s \\
Acceleration clip & $\|\ddot{\bm{x}}\|_{\max}$ & 1000\,m/s$^2$ \\
\bottomrule
\end{tabular}
\end{table}

\subsection{Autonomous Hugging Pipeline} 
For a comfortable hugging experience, ensuring both safety and an appropriate hugging position is essential. While our compliant RL policy enforces force limits for safe contact, achieving comfort requires adapting the hugging posture to the person’s body shape. 
To accomplish this, we first estimate the human body shape using BEDLAM\cite{bedlam}, and rescale it according to the subject’s absolute height obtained from motion capture. 
We then extract the waist position, denoted as $x'$, as the target contact point.

Next, we optimize the default upper-body motion of G1 so that selected robot links reach the SMPL-derived waist targets while the torso stays properly oriented in the horizontal plane. We optimize upper-body joint angles $\mathbf{q}$ and a planar floating base $\mathbf{r}=(x,y,\psi)$ with fixed height $z=z_0$. Let $\mathbf{p}_\ell(\mathbf{q},\mathbf{r})$ be the forward-kinematics position of link $\ell$, $\{\mathbf{b}_k\}$ the target points on the waist, and $\Pi_{xy}$ the $xy$-projection. The objective is
\[
\begin{aligned}
\min_{\mathbf{q},\,\mathbf{r}}\;\;
& \sum_{(\ell,k)\in\mathcal{S}} w_{\ell k}\;
   \big\|\mathbf{p}_\ell(\mathbf{q},\mathbf{r})-\mathbf{b}_k\big\|^2  \\
&\quad +\; w_t\,
   \big\|\Pi_{xy}\!\big(\mathbf{p}_{\text{torso}}(\mathbf{q},\mathbf{r})
        + \delta\,\mathbf{f}(\psi)\big)
        - \Pi_{xy}(\mathbf{b}_{\text{front}})\big\|^2 \\
&\quad +\; \lambda_{\text{reg}}\,
   \|\mathbf{q}-\mathbf{q}_0\|^2 \, .
\end{aligned}
\]
Here $\mathcal{S}$ collects the link–target pairs (e.g., hands to back-waist, elbows to opposite-side waist), $w_{\ell k}$ and $w_t$ weight their relative importance, $\delta\!\approx\!5$ cm is a small forward offset for the torso, and $\mathbf{f}(\psi)=[\cos\psi,\sin\psi,0]^\top$ denotes the heading. The regularizer $\|\mathbf{q}-\mathbf{q}_0\|^2$ keeps the solution close to a neutral upper-body pose. The optimized motion sequence is then updated as a personalized reference motion for the specific individual.

After obtaining the target posture and contact locations, the robot must first stand in the proper place. We train a locomotion policy that get the robot–human relative pose from motion-capture markers and directly commands joint targets to walk to a stance directly in front of the person, with a $10$\,cm standoff and frontal alignment. Once this condition is met, control switches to the GentleHumanoid policy to execute the hug.

\subsection{Video to Humanoid}
We use a phone to record monocular RGB videos, and apply PromptHMR~\cite{wang2025prompthmr} to estimate the corresponding human motion as an SMPL-X motion sequence. 
The estimated motion is then retargeted to the G1 humanoid using GMR. Finally, we execute the retargeted motion using our trained policy. 
As shown in the supplementary video, our method remains robust and compliant even when the estimated reference motions are noisy (e.g., with foot skating). 
It successfully handles interactions with various objects such as pillows, balloons, and baskets of different sizes and deformabilities.

\end{document}